\documentclass[sigplan,screen]{acmart}

\AtBeginDocument{%
  \providecommand\BibTeX{{%
    \normalfont B\kern-0.5em{\scshape i\kern-0.25em b}\kern-0.8em\TeX}}}

\setcopyright{acmcopyright}
\copyrightyear{2022}
\acmYear{2022}
\acmDOI{XXXXXXX.XXXXXXX}

\acmConference[ISIR-eCom 2022 @ WSDM-2022] {International Workshop on Interactive and Scalable Information Retrieval methods for eCommerce (ISIR-eCom) 2022 @ WSDM-2022, https://isir-ecom.github.io/}{Feb 25, 2022}{Phoenix, AZ}

\usepackage{tabularx}
\usepackage{subfloat}
\usepackage{subfig}
\newcommand{\name}{{\sc ORDSIM}}

\begin{document}

\title{ORDSIM: Ordinal Regression for E-Commerce Query Similarity Prediction}

\author{Md. Ahsanul Kabir}
\email{mdkabir@iu.edu}
\affiliation{%
  \institution{Indiana University-Purdue University Indianapolis}
  \city{Indianapolis}
  \state{Indiana}
  \country{USA}
}

\author{Mohammad Al Hasan}
\email{alhasan@iupui.edu}
\affiliation{%
  \institution{Indiana University-Purdue University Indianapolis}
  \city{Indianapolis}
  \state{Indiana}
  \country{USA}
}

\author{Aritra Mandal}
\email{arimandal@ebay.com}
\affiliation{%
  \institution{eBay Inc.}
  \city{San Jose}
  \state{California}
  \country{USA}
}

\author{Daniel Tunkelang}
\email{dtunkelang@ebay.com}
\affiliation{%
  \institution{eBay Inc.}
  \city{San Jose}
  \state{California}
  \country{USA}
}

\author{Zhe Wu}
\email{zwu1@ebay.com}
\affiliation{%
  \institution{eBay Inc.}
  \city{San Jose}
  \state{California}
  \country{USA}
}

\renewcommand{\shortauthors}{kabir, et al.}

\begin{abstract}

Query similarity prediction task is generally solved by regression based models with square loss. Such
a model is agnostic of absolute similarity values and it penalizes the regression error at all ranges of 
similarity values at the same scale. However, to boost e-commerce platform's monetization, it is 
important to predict high-level similarity more accurately than low-level similarity, as highly similar 
queries retrieves items according to user-intents, whereas moderately similar item retrieves related items, 
which may not lead to a purchase. 
Regression models fail to customize its loss function to concentrate around the high-similarity band, resulting poor performance
in query similarity prediction task. We address the above challenge by considering the query prediction as 
an ordinal regression problem, and thereby propose a model, ORDSIM (ORDinal Regression for SIMilarity 
Prediction). ORDSIM exploits variable-width buckets to model ordinal loss, which penalizes errors in
high-level similarity harshly, and thus enable the regression model to obtain better prediction results for 
high similarity values. We evaluate ORDSIM on a dataset of over 10 millions e-commerce queries from eBay 
platform and show that ORDSIM achieves substantially smaller prediction error compared to the competing 
regression methods on this dataset. 
\end{abstract}

\maketitle

\section{Introduction}

For e-commerce platforms, like eBay, Amazon, Alibaba, etc.\ finding queries, which
are similar to a user query is a fundamental task. Solving this task benefits various
platform features, such as  ``related searches'', ``query expansion'',  ``recovery
from zero recall'', ``auto complete'', and ``improve spelling''. These features 
enable extending the scope of search, thus improving the probability of retrieving
all relevant items for a given query. Expanding the scope of a search is particularly
important for queries, for which the user query alone does not reflect buyers' intent
effectively---also needed for queries for which  the word tokens in the query 
have poor match with the word tokens of the most relevant items. Traditional information
retrieval based document similarity prediction methods generally preform poorly for 
the case of e-commerce  queries because of the facts that queries are often shorter, 
strongly sensitive to the context, and lack syntactic structure. 

An e-commerce platform can use different cues for computing similarities between 
two queries. The obvious cue is word token based similarity, i.e., two similar
queries would share substantial fractions of the word tokens. For
example, the following four queries: ``Dell xps Laptop", ``Dell xps 15", ``Dell 15 
laptop", ``Dell 15 inch laptop", all have high word level similarity and all of 
these queries are from a buyer, whose intention is to buy a ``Dell laptop". 
Token level similarity is easy to compute as the queries generally differ
only in stemming, word order, addition/omission of words, etc. Also in
terms of query similarity, token based similarity yields high precision results.
Finally, token based similarity is simple to understand, and easy to explain.
However there are some downside of token level similarity. Often, there are
false positives; for instance, ``kiss'' is not ``kisses'', ``dress shirt'' is not
the same as ``shirt dress''. So, the similarity model requires to have guardrails 
to avoid embarrassing mistakes. Another downside is that token level similarity 
have limited coverage, and extending it significantly increases the risk of
generating false positive results.

Besides token level similarity, there are other indirect measures which can
establish the fact that two queries are similar. The mostly used e-commerce
platforms work with session-based similarity. Generally, if a committed buyer uses two
queries in the same search session, then with high likelihood these two queries
are similar. Also, queries that are semantically related are similar. For
example, one query can be a generalization or specialization of another query;
both the queries can be for items which are replaceable (different brand, size);
they can be for fitting, matching, complementary, accessory, or bundle  items. 
Two queries can also be similar if they belongs to the same node in the item 
taxonomy tree.
Note that, the different similarity cues may be indicator of different levels of
similarity. Token-based similarity, aspect-based generalization, or 
specialization generally 
indicate high degree of similarity, as such, the query pair would be intended for
identical or replaceable item. On the other hand, similarity derived from other
cues, such as, session-based similarity or taxonomy-node similarity may not be
for identical item. For instance, ``samsung 85 inch tv'' and ``wall mount for tv''
are two queries which are similar as both belongs to electronics category, but
such similarity is relatively small as the queries are not for identical or replaceable
items. 

In e-commerce domain, queries follow a power-law distribution in terms of their
frequencies over the site. Based on this, queries can be classified into head, torso,
and tail queries. Due to the long-tail distribution, a significant number of queries
belong to tail queries. Unlike head and torso queries, query similarity models 
perform much poorly for tail queries for many reasons: first, tail queries do not 
have significance presence in search session; second, not many product titles are
associated with these queries; and most importantly, the user engagement data such as, 
click, view, and purchase rate for these queries are sparse and unreliable. So, it is
better to learn query similarity for these queries by using a supervised learning
setup. In such a setup, we train a model which takes representation of a pair of
queries and their similarity values as input such that once the model is trained,
it can be used to predict similarity between a pair of tail queries for which
direct computation of query similarity is not accurate. We call this task query similarity
prediction, which is the focus of this paper. 

Given that query similarity is a continuous value, supervised prediction of this value 
seems to be a  straightforward regression task. However, for better interpretation, the 
similarity value should be a closed range, such as, between 0 and 1. This makes it a
regression task with additional constraint that the target value should be between 
0 and 1. Another important requirement is that we want to predict high-level 
similarity (say similarity above 0.90) more accurately than low-level similarity (say between
0.5-0.9). This is due to the fact that correct usages of highly similar query
helps buyers to retrieve her desired item and significantly boost platform's 
monetization. This is similar to the fact that in information retrieval 
accurate ranking of search results at the head is much more important than 
the same at the torso or tail. Unfortunately, a regression task with
square loss is agnostic of absolute similarity values  and such a loss penalizes the error
at all ranges of similarity values at the same scale. So, alternate models are 
needed which can prioritize the accurate similarity prediction of high similarity
values than low similarity values. 

For supervised prediction of query similarity, the way the queries are represented
is an important consideration. While existing works uses off-the-shelf text embedding 
methods~\cite{w2vec:Embedding,Logeswaran2018,pennington-etal-2014-glove,USE2018,akbik-etal-2019-flair,fasttext:Embedding,bert:Embedding}, 
we explore different text embedding options to identify better query 
representation vectors for the task of query similarity prediction for e-commerce data.

In this work, we solve supervised query similarity prediction task by designing an
ordinal regression model using neural networks. Our solution puts strong emphasis on the
high similarity value ranges than low similarity value ranges, which serves a critical
need in the production environment. Note that, although introducing more similarity buckets can make the query similarity prediction task a ranking problem, we do not intend to solve a ranking problem since within a similarity bucket the ranking among queries based on similarity scores is more often noise. We also develop innovative query representation
vectors using spherical text embedding~\cite{spherical:Embedding}, which shows improved performance over
well-known BERT~\cite{bert:Embedding} -based text embedding. Using $\mathcal{U}$, an eBay dataset with around 10 million queries, \name\ shows substantial improvement over the existing state-of-the-art. 

\section{Related Works} \label{rel}
Document similarity prediction is a well studied task in the information retrieval domain ~\cite{Sun2021,Devlin2019BERTPO,siamese2015,sim2017,peng-sim-2020}. However, most of the
works in this domain predict similarity for web documents or traditional textual literature.
These works are poor fit for e-commerce queries as e-commerce queries are short, and lacks
syntactic structure. Among the existing works, Fuchs et al.~\cite{Fuchs2020IntentDrivenSI} 
focuses on similarity on e-commerce product listings. They use pairs of listing titles and 
their matching search queries, and leverage a contextualized character language model, L2Q 
works as a bidirectional recurrent neural network to produce token importance weights. They demonstrate
that plugging these weights into relatively straightforward listing similarity methods significantly improve the similarity results.
A plenty of researches are performed for query suggestion which are generic and more applicable 
for search engine queries~\cite{Qi2016LocationAK,PARIKH2013493}. 
There exists an unsupervised approach for query suggestion in e-commerce domain where the query 
similarity scores are calculated based on popularity and purchase-efficiency of queries~\cite{hasan2011}. Other works focus on query re-writing among which Shing et al. provide an unsupervised approach~\cite{Singh-2012}, and Hirsch et al.~\cite{Hirsch2020} focus on whether a user will reformulate a query, Aritra et al.~\cite{Mandal2019QueryRU} generate synonyms for query rewriting, and Xiao et al.~\cite{Xiao2019} focuses on dataset generation to reduce the gap between user query and product listings. 


We adopted spherical text embedding based query representation in this work.
In earlier works, representation learning in e-commerce domain is performed using feature 
engineering~\cite{Baeza2007,Sondhi2018}. Traditional embedding approaches~\cite{Fuchs2020IntentDrivenSI} or 
joint embedding of queries and items into a common vector space~\cite{Ahmadvand2021DeepCATDC} are also considered. 
For text representation, a plenty of research have been performed which include Word2Vec~\cite{word2vec-2013}, GloVe~\cite{pennington-etal-2014-glove}, FastText~\cite{fasttext:Embedding}, ELMo~\cite{peter-elmo-2018}, Flair~\cite{akbik-etal-2019-flair},  TSDAE~\cite{Wang2021}, InferSent~\cite{Conneau2017}, QuickThought vectors~\cite{Logeswaran2018}, Universal Sentence Encoder~\cite{USE2018}  and perhaps most notably BERT~\cite{bert:Embedding}. 
Recently, spherical text embedding~\cite{spherical:Embedding} has been considered for text embedding, in which
a text snippet is embedded on the surface of a d-dimensional unit sphere. In this work, we show that for 
short text, like e-commerce queries, spherical embedding performs better that BERT-based embedding.

A plenty of research works are performed in ordinal regression~\cite{Chu2007,Zhu2021Ordinal,Berg2020Ordinal,Hiba2021,Kleindessner2021,Gu2020,liu2021}. Among these,
Wenzhi et al.~\cite{coral-ordinal} uses a neural network architecture with a non-traditional loss function which is particularly suited for the ordinal
regression task. CORAL model can perform ordinal regression directly\cite{coral-ordinal} and works with the last layer of a neural network like \name. Instead of one softmax unit at the
output layer, $K-1$ sigmoid classifiers are designed for ordinal regression where $K$ is the total number of ordinal labels. Additionally, Jason et al.~\cite{Rennie2005} compiled a list of loss functions for ordinal regression and discussed their relative strength
and weakness.

\section{Problem Formulation}\label{definition}
Let $\mathcal{D} = {\{x_i, y_i\}}_{i=1}^N $ be a training dataset comprised of $N$ data instances where $i$'th training sample $x_i$ contains two e-commerce queries $q_1$, and $q_2$, i.e. $x_i = (q_1, q_2)$. 
Since most e-commerce platforms have tool-set for automatically allocate the query to one or
multiple taxonomy nodes, we assume that $c_1$, and $c_2$ are associated taxonomy nodes of query 
$q_1$, and $q_2$, respectively. Furthermore, $y_i$ is the similarity value of the two queries in $x_i$,
where $0 < y_i \le 1$. Supervised query similarity learning task is to train a model so that the model
is able to predict the similarity value between two queries for which the similarity value is unknown.

\begin{figure*}[t!]%
    \centering
    \subfloat[\centering The frequency distribution of the queries based on similarity values]{{\includegraphics[width=7.6cm, height=4.2cm]{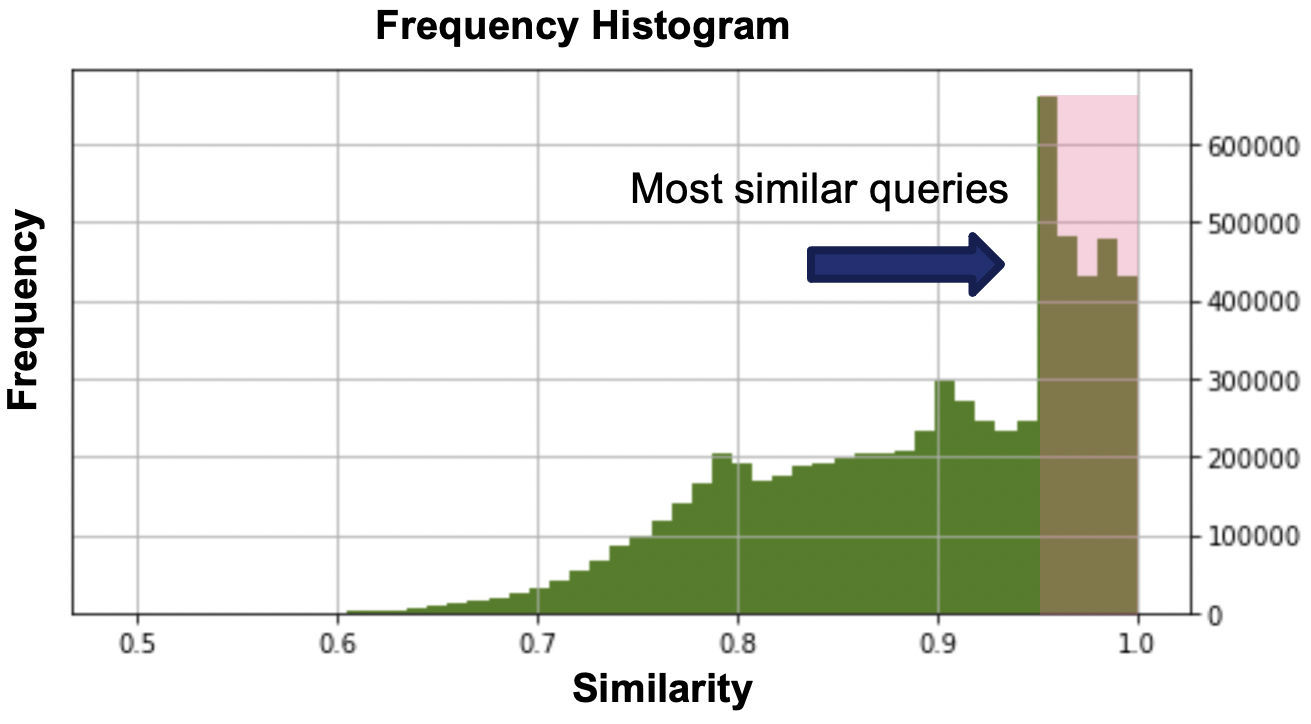} }}%
    \subfloat[\centering The cumulative distribution of the queries based on similarity values]{{\includegraphics[width=7.6cm, height=4.2cm]{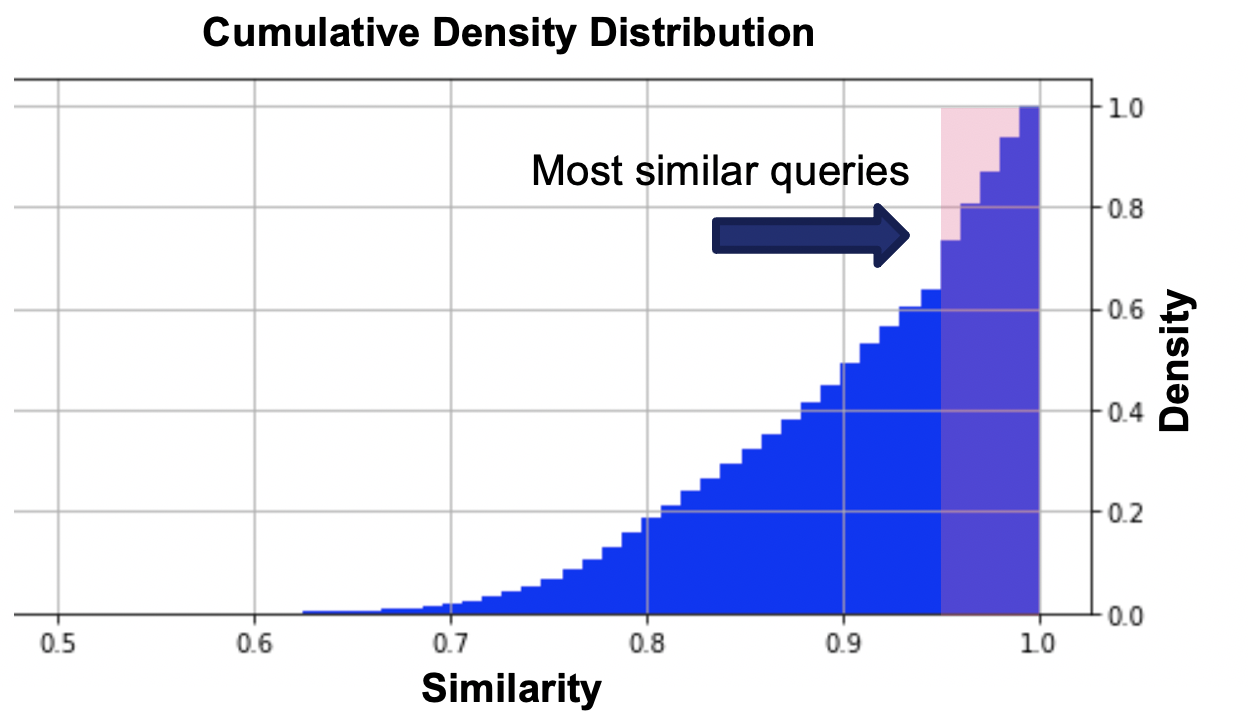} }}%
    \caption{Distribution of the similarity values in the original dataset for query similarity prediction}%
    \label{fig:distribution}%
\end{figure*}

For training we use $\mathcal{D}$, a subset of $\mathcal{U}$ which contains 6 million query-pairs and their similarity
values as obtained from Cosine similarity of the embedding vectors of two queries in a pair. Distribution
of similarity values in $\mathcal{U}$ is shown in Figure~\ref{fig:distribution}. Left side of Figure
\ref{fig:distribution} shows the distribution of the similarity values, and the right side of the 
figure shows the cumulative distribution of the same data. As we can see from these charts, the
distribution of similarity values are skewed towards higher similarity values. In fact, most similarity 
values are higher than 0.5. About 36\% of similarity values in the dataset falls between
[0.95, 1]. This above behavior is expected because $q1$ and $q2$ are not uniformly chosen
random query pair, rather they are related queries through product listing, 
user session, taxonomy node mapping, etc. The justification of using related query-pairs for $\mathcal{U}$ is as follows: If two queries are not related, predicting their similarity is not important 
because they are usually not even candidates for query reformulation or query expansion/modification.
On the other hand if multiple queries are highly similar to a given query, most of them are 
candidates for query reformulation or query expansion/modification and a correct ordering among the candidates
is critical. Hence, the dataset only contains query-pairs
that are known to be similar with a high similarity value and our task is to correctly rank the similar 
queries of a given query by predicting the correct similarity values using a supervised model.

\begin{figure}
    \centering
    \includegraphics[width=1\linewidth]{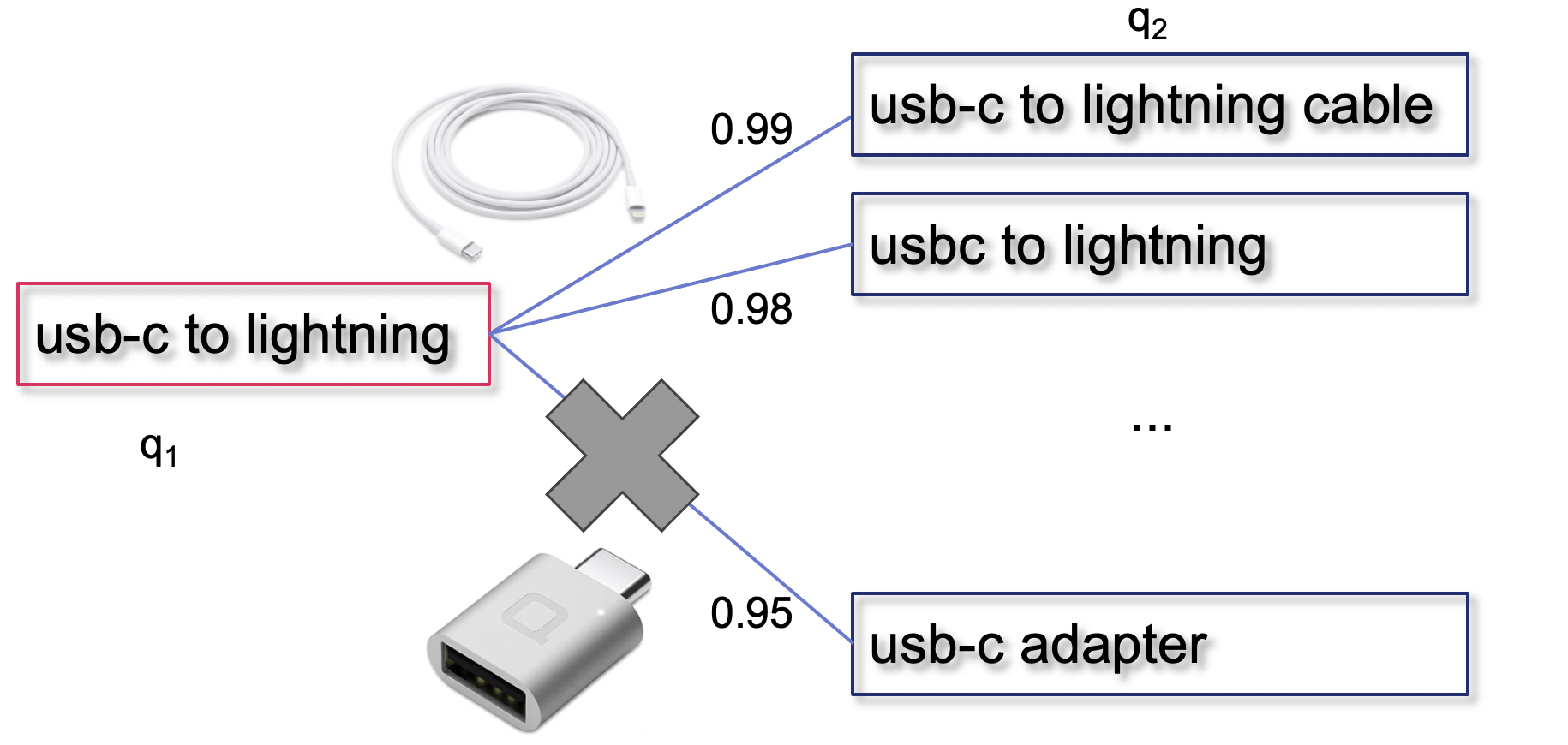}
    \caption{Implication of range of query similarity values}
    \label{fig:example}
\end{figure}

In Figure~\ref{fig:example}, we provide a real-life example from query set. Say, a query is ``usb-c to lightning"; from our query embedding vectors, three most similar
queries to this query are: ``usb-c to lightning cable", ``usbc to lightning", 
and ``usb-c adapter" with similarity values 0.99, 0.98, and 0.95. Among these three
queries, ``usb-c adapter" is actually a different object even though the similarity
value is a whopping 0.95! A correct ordering of the similar queries can help filter out false-positive similar queries with high similarity values. This example
demonstrates that correct similarity prediction at a high similarity value is much more 
important than the same at a moderate or a low similarity value. One can also formulate 
this problem as a ranking problem, where the similarity rank orders at the head of the list is much
more important than the similarity rank order at the torso or tail. 
Nevertheless, we do not
want to solve this problem as a full ranking problem as the ranking model has higher complexity
due to larger number of rank order constraints; specifically, for a given query if we consider 
$k$ similar candidate queries, we have $\omega (k^2)$ binary rank order constraints. 

A key challenge for solving this problem using regression is that it penalizes the prediction
error at all ranges of similarity values at the same scale. However, for our dataset, $\mathcal{U}$ very high 
similarity value represents identical query, moderate similarity value represents replaceable,
alternate, or generalization/specialization of a query, and small similarity value represents query 
for other related items. Among these correctly predicting the similarity between highly similar 
queries is much more important than correctly predicting the similarity between moderate or small similarity
queries. For instance, predicting the similarity value of 0.97 as 0.93 is much worse than predicting 0.70 as
0.74, although both have a deviation of 0.04.

To prioritize correct prediction of highly similar queries, we model the supervised query similarity
prediction task as ordinal regression with ordinal value from non-uniform bucket size. At the high
similarity value range, the bucket width is small, so a small deviation would cause an ordinal value 
error. On the other hand the bucket width is large at small similarity value. For instance, for the 
similarity data shown in the graph in Figure~\ref{fig:distribution} (left),

For example, we can bucket the queries into five classes with the following similarity range (bucket) to ordinal label
mapping: $(0-0.82]\to 0, (0.82-0.9]\to 1, (0.9-0.95]\to 2, (0.95-0.97]\to 3$, and $(0.97-1]\to 4$. Obviously
higher ordinal label denotes higher level of similarity. Using the above bucketing, for our query similarity
dataset, $\mathcal{U}$ each bucket contains approximately 1.2-1.6 million queries but due to right skewed distribution, ordinal
label representing high similarity value has a very small similarity value range, on the other hand 
ordinal value with small similarity value has a very large range. For instance, If there are two similarity
values, say  0.4, and 0.65, both similarity are considered to be the same by ordinal label. On the
other hand, similarity value above 0.97 is considered higher similarity than the similarity value of 0.95,
as the former has an ordinal label of 4 and the latter has an ordinal label of 3. This non-uniform 
similarity width bucketing forces the model to correctly distinguish query similarity between Label 3 and
Label 4; although both represent very high similarity (above 0.95), Label 4 may represents two queries which 
retrieves identical objects, on the other hand, Label 3 may not do so. We use $\mathcal{H}$ to denote the
similarity value to ordinal label function.

\subsection{Mapping Similarity Value to Ordinal Label} \label{ranges} Finding correct Ordinal buckets for 
similarity values in e-commerce queries requires experts' observations and expertise. Additionally, the method of 
similarity calculation and distribution of the similarity values are also important. In this work the similarity 
values are Cosine similarities between the vector representations of corresponding query pairs.

A carefully designed map function $\mathcal{H}$ transforms $y_i \rightarrow l_i$ where $l_i$ is the ordinal
label of $y_i$, and $l_i$ $\epsilon$ $\{r_j\}_{j=1}^K$ 
such that $r_1 \prec r_2 \prec ... \prec r_K$. The goal is to represent $x_i$ in such a way that an Ordinal classification task captures the best ranking 
function 

$f : X \rightarrow l$ which minimizes a loss function, $\mathcal{L}$.

\begin{figure*}
    \centering
    \includegraphics[width=0.8\linewidth]{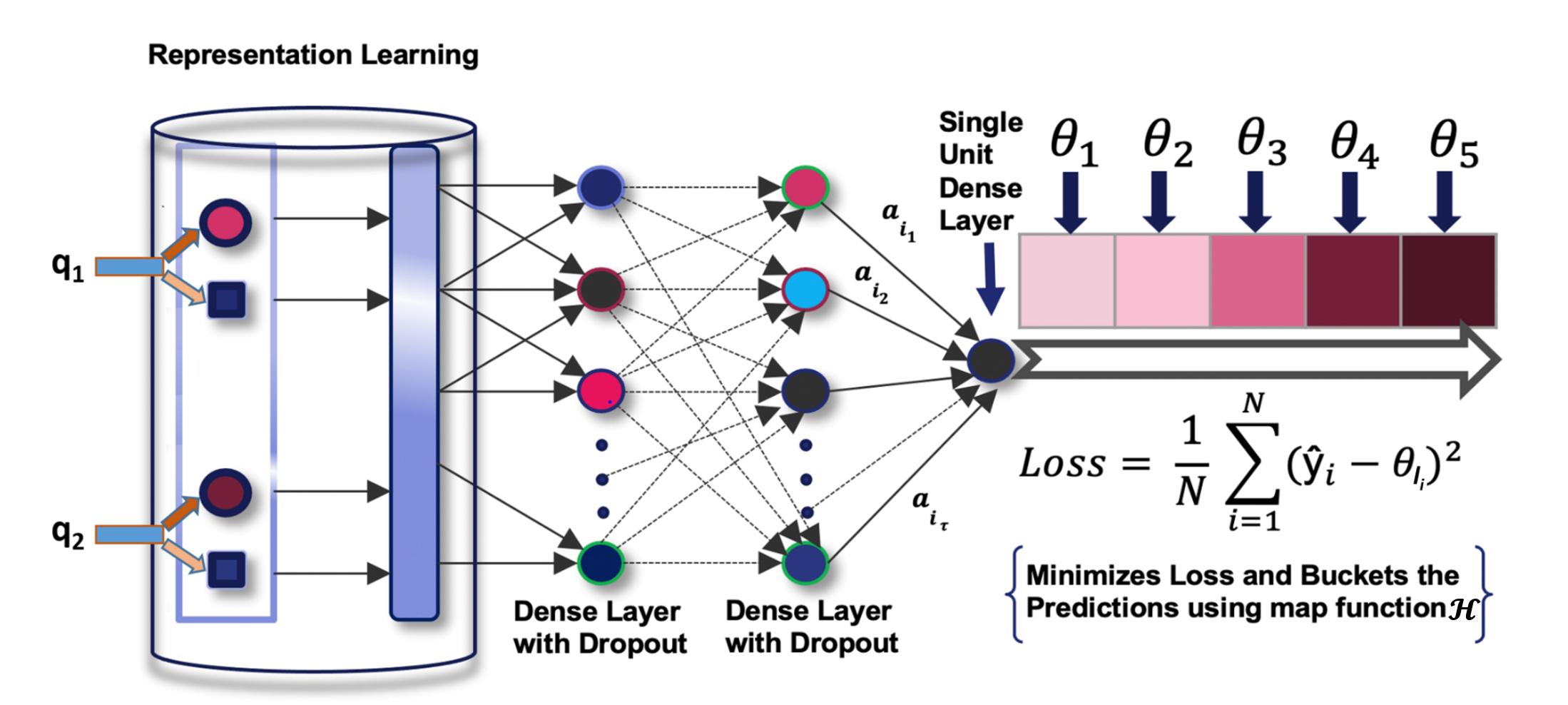}
    \caption{Illustration of \name\ model to perform query similarity prediction task. $q_1$, and $q_2$ are passed for representation learning followed by a multilayer perceptron and ATMSEL loss function in the output layer to perform ordinal regression.}
    \label{fig:model}
\end{figure*}

\subsection{Could we perform multi-class classification?} 
If we consider the ordinal label as one class in a set of classes, then this task can also be solved as multi-class
classification. However, for multi-class classification on such data, the categorical cross-entropy loss for wrong
prediction of a single instance is constant and is oblivious of the magnitude of ordinal loss. So, on such data,
multi-class classification is a poor fit.

Say, $\mathcal{P}$ be a loss matrix of shape $K*K$ where $\mathcal{P}_{lr_k}$ is the loss for predicting 
the label $l$ as $r_k$. Note that for both classification and ordinal regression, the loss $\mathcal{P}_{ll} = 0$ and $\mathcal{P}_{lr_k} > 0$ when $l \neq r_k$. For classification, misclassification loss, $\mathcal{P}_{lr_{k}}$ is fixed $\forall k \in [1:K]$, where $l \neq r_k$. But for ordinal regression, the loss varies based on the magnitude 
of ordinal gap between the actual label and the predicted label; for
instance, one can define the ordinal loss $\mathcal{P}_{lr_{k}} = |l - r_k|$ as both $l$ and $r_k$ are numeric 
values. Hence, ordinal regression is a preferred task over the multi-class classification, as the former reflects
the extent of loss more accurately.

\subsection{Methodology}
\name\ uses a neural network based model for performing ordinal regression for query similarity prediction. 
Specifically to build \name, we design the output layer of a neural network so that it can adopt an 
ordinal loss function. In this way, \name\ is architecture independent and any neural network model can be made 
an ordinal regression model only by applying \name's output layer.
Below, we discuss \name's loss function which the model optimizes.

\subsubsection{Loss Function} \label{loss}
We design a loss function for \name, which we refer as ``All Threshold Mean Squared Error Loss function'', in short, ATMSEL.
For a multi-layer perceptron, for an instance $(x_i, y_i)$, say $l_i$ is the  mapped ordinal label, after using
mapping function $\mathcal{H}$ on $y_i$. ATMSEL penalizes a predicted instance based on how far the predicted 
score $(\hat{y}_i)$ is from the mid-range of $l_i$ bucket. If $\theta_{l_i}$ is the mid-range of ordinal bucket $l_i$, 
the ATMSEL loss function can be represented by the following equation:

\begin{equation}
\label{eq:loss}
    ATMSEL = \frac{1}{N}\sum_{i=1}^{N} (\hat{y}_i - \theta_{l_i}) ^ 2 
\end{equation}

During training,  the loss function ATMSEL is applied at the output layer of the perceptron over a mini-batch and the model 
is trained by using backpropagation. In Figure \ref{fig:model}, towards the right side, we show the mid-range 
of all $K$ ordinal buckets ($K=5$ ordinal bucket are shown in the figure). Note that, both $y_i$ and $\hat{y}_i$ are similarity value between 0 and 1. Their ordinal representation is only utilized
in the loss function in Equation~\ref{eq:loss}. Also note that in our current work, we fixed the left and right interval of 
similarity values for an  ordinal label by observing the distribution of similarity values, but one can also adopt a more sophisticated methods for obtaining the similarity interval for each ordinal label.

To train the model let $x_i = $ $(q_1, q_2)$ be the i'th training sample of $\mathcal{D}$. The representation vector of $x_i$ is the input of a neural network which is connected to two dense layers. Say, for the data instance $x_i$, $a_{i_1}$, $a_{i_2}$, ... ,$a_{i_\tau}$ are the output values of the penultimate layer of the network. If, the weights associated with these values
are $w_1$, $w_2$, ... ,$w_\tau$ respectively. the predicted similarity value, $\hat{y}_i$, is calculated by the following equation.

\begin{equation}
    \hat{y}_i = \sum_{j=1}^{\tau} a_{i_j} * w_j
\end{equation}
 
For instance, say an actual similarity value of two queries in the train dataset is 0.95 which resides between (0.9-0.95] bucket, and $\theta_{l_i} = 0.925$, and the corresponding predicted similarity value $\hat{y}_i$ is 0.901. As ATMSEL penalizes based on deviation from $\theta_{l_i}$, the error for this instance is $(0.901 - 0.925)^2$. Furthermore, while ATMSEL is minimized, the predicted values tend to get centralized to the median of the corresponding buckets. That is why ATMSEL can capture the query similarity buckets better than other methods. 
\subsubsection{Model Architecture} \label{model}
We illustrate the architecture of the deep learning model in Figure~\ref{fig:model}.
The main model is kept same for both CORAL and \name. To describe the model, recall $x_i = $ $(q_1, q_2)$ where $x_i \in \mathcal{D}$. Both $q_1$ and $q_2$ contain query text and category path text information which are shown by two circles and two squares respectively as described in the left side of Figure~\ref{fig:model}. $q_1$ and $q_2$ are then passed for representation learning. The embedding of $q_1$ and $q_2$ are the inputs of the main model. Followed by the embedding layer, two hidden layers are added with $N_{\kappa}$ and $N_{\tau}$ neurons with dropout probabilities of $p_1$ and $p_2$ respectively. For both hidden layers we use $ReLU$ activation function. The last layer differ between CORAL and \name. While we have already demonstrated ATMSEL loss function for \name, the CORAL model introduces $K-1$ sigmoid classifiers to perform the ordinal regression task.

\subsection{Evaluation Metric} 
The evaluation metric for 
ordinal regression should measure the extent by which the predicted ordinal labels deviate from the actual ordinal labels. 
However, neither the actual similarities ($y_i$s), nor the predicted similarities ($\hat{y}_i$s) are ordinal. So, we first 
map both $y_i$ and $\hat{y}_i$ to ordinal values using the map function $\mathcal{H}$. Say, the corresponding ordinal label of both $y_i$ and $\hat{y}_i$ are $l_i$ and $\hat{l}_i$, respectively. Then the evaluation metric Mean Average Label Error (MALE), 
is defined by the following equation:

\begin{equation}
    MALE = \frac{1}{N}\sum_{i=1}^{N} |\hat{l}_i - l_i| 
\end{equation}
MALE denotes the extent by which a predicted ordinal label  deviates from its actual ordinal label.

\subsection{Representation Learning} \label{rep}
For applying the neural network model for the query similarity prediction task, the queries need to be represented as
vectors in real space such that similar queries are embedded in close proximity which enables high query-query similarity
prediction. We perform this through representation learning. Given the dataset $\mathcal{D} = \{x_i, y_i\}_1^N$, 
$x_i$ = ($q_1, q_2$) consists of a pair of e-commerce queries. For both the datasets $\mathcal{D}$ and $\mathcal{U}$ we also have a product taxonomy node (aka
category node) associated with each query, so we leverage the category information to learn better representation of the queries.
One can simply ignore the category as part of query representation if such information is not available.
Both queries, and categories are text data, but the categories are also organized in a tree structure which we refer as category tree. For a query, along with the query text, the path-label of the nodes along the path from the root to the associated 
category node in the category tree is used for query representation.

To illustrate let a query be ``t shirts for men pack"; for $\mathcal{D}$, the category for this query is \textit{Shirt}. In
the category tree  the path to the \textit{Shirt} node has the following labels: \textit{Clothing->Men's Clothing->Shirt}. 
The text tokens in this path-label, along with the text token from the queries are embedded in a vector space to obtain
for embedding of the query. Many text embedding methods, such as,
BERT~\cite{bert:Embedding}, Word2Vec~\cite{w2vec:Embedding}, FastText~\cite{fasttext:Embedding} are popular in the information
retrieval domain, but they are more suited for traditional text documents. E-commerce queries are different from traditional
text documents as the former lacks proper syntactic structure.
Additionally, most of the embedding methods like BERT~\cite{bert:Embedding}, Word2Vec~\cite{w2vec:Embedding}, FastText~\cite{fasttext:Embedding} depend on training a large corpus such as Wikipedia, Google and the sentences in these datasets are different than e-commerce queries as most of the sentences in the datasets are complete and natural. This motivates us to build an alternative approach for query embedding, after modifying eBERT using spherical query embedding. Below we provide more details about these tools.

\subsubsection{BERT and eBERT}
Bidirectional Encoder Representations from Transformers (BERT) is proposed by researchers from Google~\cite{bert:Embedding} which is not trained on any specific downstream task but instead on a more generic task called Masked Language Modeling. The idea is to leverage huge amounts of unlabeled data to pre-train a model on language modeling. Predicting the next word(s) given a context already requires understanding language to some extent. Next, this pre-trained model can be fine-tuned to solve different kinds of NLP tasks by adding a task specific layer which maps the contextualized token embeddings into the desired output function.

eBERT\footnote{https://www.enterpriseai.news/2021/09/15/heres-how-ebay-is-using-optimization-techniques-to-scale-ai-for-its-recommendation-systems/} is an e-commerce specific version of the BERT model. Along with the Wikipedia corpus, 1 billion latest unique item titles are collected to train the model. The eBERT model can represent the e-commerce terms better than just using only Wikipedia corpus.  In this work, to train the models, we use eBERT to embed both query and category text.
We keep the embedding dimension of eBERT similar to BERT which is 768.

\subsubsection{Spherical Text Embedding}
Directional similarity is often more effective in tasks, such as, word similarity and document clustering. When textual units
are embedded in the Euclidean space, two textual units may have zero directional distance, yet they are far from each other
by Euclidean distance. This is not desirable when we are trying to predict the query similarity values, which are between
0 and 1, and was computed by directional similarity (Cosine) of two vectors. To overcome this, spherical text embedding 
\cite{spherical:Embedding} has been proposed, which embed the textual units on the surface of a unit d-sphere 
(d is the dimension). To learn embedding on unit d-sphere, an efficient optimization algorithm is proposed with convergence guarantee based on Riemannian optimization. Spherical text embedding shown to be highly effective on various text
embedding tasks, including word similarity and document clustering. For embedding e-commerce query text, we also choose
spherical text embedding as a candidate. To build the corpus of spherical text embedding, besides Wikipedia sentences, we introduce all of the 10 million e-commerce queries in the dataset, $\mathcal{U}$. We choose the dimensionality of the sphere as 100 as
recommended by the original authors. We use each word as a textual unit for embedding task and then apply mean pooling over 
the words of a query to obtain query embedding vectors.

\subsubsection{Poincaré Embedding} While BERT, eBERT, and spherical text embedding embed the category names as text data, the hierarchy information of the category nodes remains untapped by the embedding methods. Since the categories 
form a tree, and Euclidean space is not a good fit for representing a tree structure, in recent works many embedding 
methods are proposed which embed a tree-like data in hyperbolic space; for instance, Nickel et al.\cite{nickel2017poincare}
propose to embed a tree in a n-dimensional Poincaré ball. The neighboring vertices in the tree are expected to get closer 
in the hyperbolic space in terms of Poincaré distance. The loss function is optimized by Riemannian optimization. 

In this work we embed the category tree described in section~\ref{rep} with Poincaré embedding, and embedding of the category is used to perform different experiments. Note that, we do not have query embedding with Poincaré Embedding as the queries do not form any tree-like structure. To embed the category tree with Poincaré model we tune the embedding dimensions in range [5,120] with an interval of 5. Furthermore, the epochs are tuned in range [50,1000] with an interval of 50. The best embedding is picked in terms of best Mean Average Precision (MAP) which we get for dimension = 10, and epoch = 500.

\subsection{Embedding of Query Pairs for Similarity Prediction}
Recall $\mathcal{D} = \{x_i, y_i\}_1^N$ where
$x_i$ = ($q_1, q_2$). Both $q_1$ and $q_2$ contain query text, but category information may or may not be added with $q_1$ and $q_2$. If category information is excluded, we simply embed query text using mean pooling of word token representations. For the other two way of representations, category information is added. To embed query text we use both eBERT and spherical text embedding. While for category embedding along with eBERT and spherical text embedding, we use Poincaré model as well. To embed category path with eBERT and spherical text embedding mean pooling of the whole category path word tokens representations is used. In contrast, only the last category in the category path, which is the actual category is embedded with Poincaré model to represent category embedding.

\section{Experiments and Results} 

We perform comprehensive experiment to show the effectiveness of \name\ for query similarity prediction. Besides, we also
perform experiments to show how different embedding choices affect the performance of query similarity prediction task.
Below we first discuss the dataset, and competing methods, followed by experimental results of the experiments.

\begin{figure}
    \centering
    \includegraphics[width=1\linewidth]{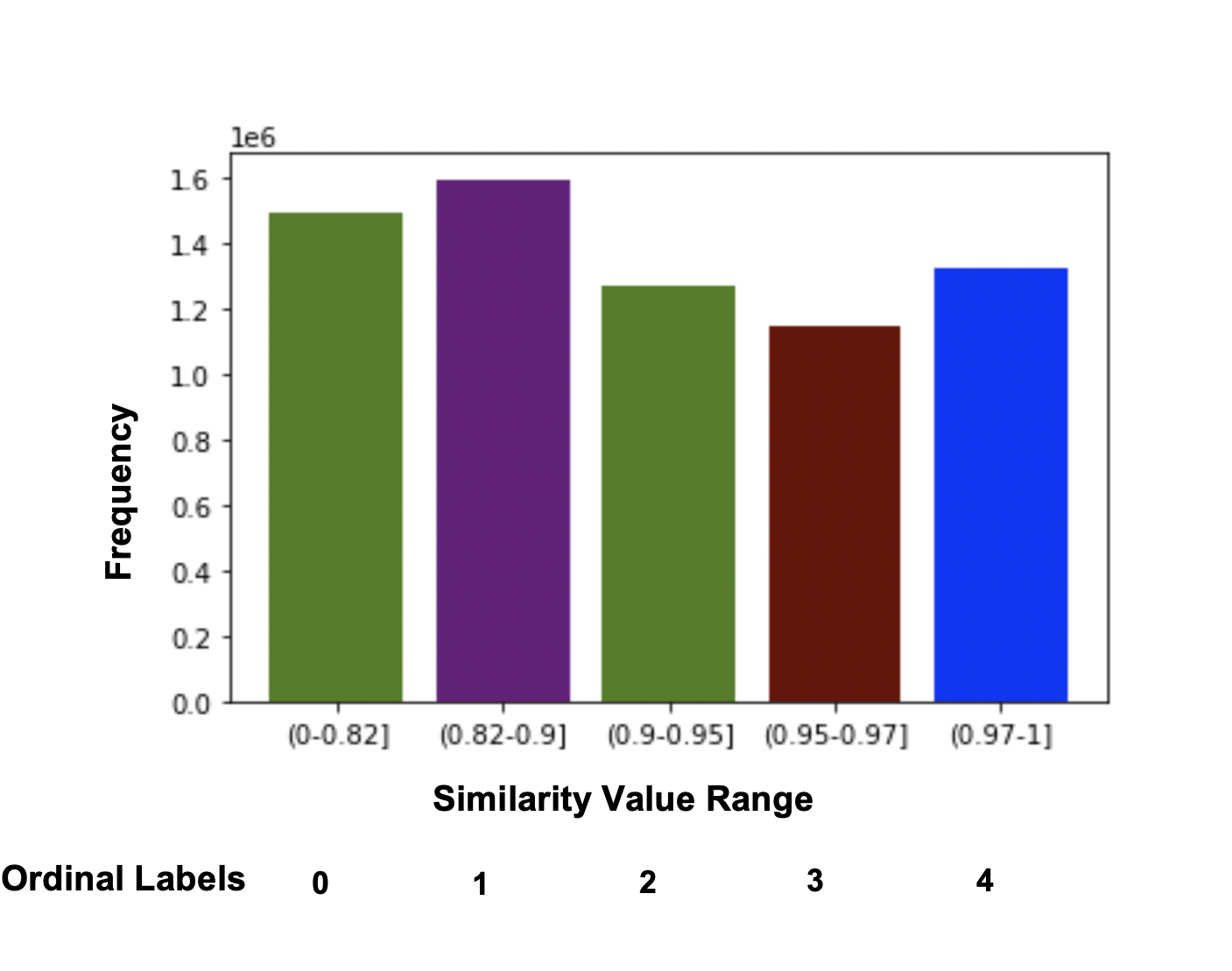}
    \caption{Frequency Distribution for Different buckets of Ordinal Labels }
    \label{fig:buckets}
\end{figure}

\subsection{Dataset} \label{dataset} 
Our dataset, $\mathcal{U}$ contains two eCommerce queries, their category nodes in terms of a prototypical e-commerce taxonomy, and the ground truth similarity values between a pair of queries. We build the ground truth similarity value between two queries by building association between two
queries through the product titles aka listings against those queries. For a given query, we first rank the listings based on
hit rate and purchase rate. Once there is a ranking of listings with respect to a given query, the corresponding query is
embedded by mean pooling of the top product titles' representations where the product titles are embedded using fasttext embedding method\cite{fasttext:Embedding}. Then the similarity values between two queries are simply Cosine similarities of 
the corresponding query pairs' representation vectors. Note that for $\mathcal{U}$, titles are missing but similarity values are provided. The goal of \name\ is to represent the queries without using product titles such that the similarity between queries can be captured. Note that it is not expected that every e-commerce query will have associated meaningful and adequate product titles. That's why the query similarity prediction task aims to capture the most similar queries irrespective of having adequate associated titles.

\begin{table*}[!t]
\caption{Comparison with Baseline Methods for all the representations}
\setlength{\tabcolsep}{12pt}
        \centering
     
                \begin{tabular}{l|c|c|c}
                    \hline
                    \hline
                    \bf Baseline Methods & \bf Query Embedding & \bf Category Embedding & \bf MALE
                    \\\hline
                    \hline
                     &   & None & 0.981 \\
                     & \textbf{eBERT} & Poincaré & 0.977 \\
                     Linear Regression &  & \textbf{eBERT} & \textbf{0.954} \\\cline{2-4}
                     &  & None & 0.968 \\
                     & \textbf{Spherical} & Poincaré & 0.953 \\
                     &  & \textbf{Spherical} & \textbf{0.923} \\
                     \hline
                     \hline

                     &   & None & 0.895 \\
                     & eBERT & Poincaré & 0.885 \\
                     CORAL-Ordinal &  & eBERT & 0.816 \\\cline{2-4}
                     &  & None & 0.851 \\
                     & \textbf{Spherical} & Poincaré & 0.843 \\
                     &  & \textbf{Spherical} & \textbf{0.781} \\
                     \hline
                     \hline
                     &   & None & 0.791 \\
                     & eBERT & Poincaré & 0.798 \\
                     \textbf{ORDSIM} &  & eBERT & 0.567 \\\cline{2-4}
                     &  & None & 0.613 \\
                     & \textbf{Spherical} & Poincaré & 0.559 \\
                     &  & \textbf{Spherical} & \textbf{0.459} \\
                     \hline
                \end{tabular}
            \label{table:results}
\end{table*}

The parent query similarity prediction dataset, $\mathcal{U}$ contains 10 million entries, from which train, test and validation datasets are sampled randomly maintaining 6:2:2 ratio. All the train, test, and validation datasets are disjoint. The frequency distribution of the similarity values of the parent dataset, $\mathcal{U}$ is shown in left side of Figure~\ref{fig:distribution}, whereas the cumulative distribution is shown on the right side. 
The distribution of similarity values is not uniform in the 0-1 range, rather it is right-skewed (similarity values are between 0.6 to 10); this is due to the fact that query-pairs in the datasets are not 
randomly chosen, rather they are chosen based on other similarity ques (such as category match, word token match, etc.). This
is done intentionally because for query similarity prediction, we want the model to be able to predict similarity and correctly
rank among the top candidates, not among arbitrary chosen queries. This also requires similarity prediction at the very high
similarity value to be very accurate, even if it comes at the expense of poor prediction performance at lower similarity value. 

Since we use want the similarity prediction at the high similarity value to be very accurate, we choose similarity
interval of the ordinal labels judiciously. By carefully observing the similarity value distribution we have chosen 
5 ordinal intervals between 0 and 4. The similarity value interval of these ordinals labels are as below:  0: (0.00 - 0.82], 
1: (0.82, 0.90], 2: (0.90, 0.95], 3: (0.95, 0.97], and 4: (0.97, 1.0]. The frequency distribution over different ordinal
labels are shown in Figure~\ref{fig:buckets}. As can be seen, the frequency of the ordinal labels are roughly uniform.

\subsection{Competing Methods} \label{sec:baselines} 
There is hardly any research which performs ordinal regression in e-commerce queries. For baselines, we choose linear regression, and CORAL\cite{coral-ordinal} which can work on any corpus. Each of the baselines is trained with different representations of queries and categories. Note that, eBERT and spherical embedding for text can embed both queries and categories but Poincaré Embedding is performed only for categories.

\subsubsection{Linear Regression} Linear regression (LR) although is not a good fit with ordinal regression, we consider LR as one of the baselines. The main reason for that is the query similarity values are well fitted with linear regression. Moreover, looking into the mean squared error (MSE) values, representation learning of the queries and categories can be validated easily. Finally, both the predicted and actual similarity values can be mapped to ordinal labels using the same map function $\mathcal{H}$. Once the ordinal labels are calculated, MALE can be calculated easily to evaluate the performance of LR.

\subsubsection{CORAL-Ordinal}
 We use CORAL-Ordinal as one of the baselines which we have already described in section~\ref{rel}. Both \name, and CORAL introduce distinguished loss functions for ordinal regression. For a fair comparison, we keep the main model fixed for both of the methods. Only the output layer is changed between these two.

\subsection{Training and Hyper-Parameters} For all the experiments, we have the previously described train, test, and validation splits.
For each of the competing methods, the train split $\mathcal{D}$ is used to train the model, the hyper-parameters of the best model are picked on the basis of the performance in the validation split, and each method is judged by the performance in the test split.

For both CORAL and \name\, we tune $N_{\kappa}$, $N_{\tau}$ for values {\{32, 64, 128, 256, 512\}}; and $p_1$ and $p_2$ are tuned for probability values [0.1-0.9] with an interval of 0.1. For both CORAL and \name, we get best performance for $N_{\kappa} = 256$, $N_{\tau} = 128$, $p_1 = 0.4$ and $p_2 = 0.1$. Number of epochs is 1000 for both methods. Both of them use early stopping with patience 30, and adam optimizer to converge. Lastly, the LR model is trained with the identical representation set up of for $x_i$ to minimize mean squared error loss. After getting the predicted similarity scores $\mathcal{H}$ is applied to both predicted and actual similarities for evaluation purpose.

\textbf{\subsection{Results}} In this subsection, we want to provide an extensive evaluation of all the learned query representations along with the baseline methods in terms of MALE. We want to show the best query representation, and overall best score by a method. Table \ref{table:results} shows the performance of all the methods, and the winning representations and method(s) are marked with bold letters.

 It is evident from the result that concatenating the embedding of the categories in the representation vector reduces MALE which proves the necessity of the category information in a query similarity task. Moreover, adding Poincaré embedding slightly reduces error than excluding the category information from the representation vector totally. Note that the learned vector in Poincaré space is different than Euclidean and Spherical embedding. Alongside the embedding space mismatch issue, for all the methods neural networks are trained in the  Euclidean space. For linear regression, queries and categories represented by spherical text embedding achieves the best MALE (0.923), which is 3\% better than Poincaré embedding. For all the methods, spherical text embedding for both query and category embedding provides the best representation of a term pair. For CORAL, MALE for the best representation is 0.781 which is 4\% better than eBERT(0.816). CORAL performs better than linear regression for every representation. However, in terms of MALE, \name\ achieves the best score (0.459) which is approximately 42\% better than the best of CORAL. Moreover, in terms of every representation \name\ performs better than any of the baselines including CORAL. This is because the CORAL method does not use one hot encoding of the labels. If one of the classifiers fails, count of consecutive ones from the left reduces which increases the MALE. Additionally, the performance in the most similar queries is affected. But for the e-commerce queries, identifying the most similar queries are very important which advocates the necessity of the proposed \name\ method.
 
 To illustrate with an example, let the original ordinal label, $r_i$ for $x_i$ is 3. The CORAL represents $r_i$ as [1,1,1,0] as $K$ = 5. If the second classifier of CORAL fails, the predicted output is [1,0,1,0]. Note that the output will be interpreted as 1, as the number of consecutive ones from the left to right is 1. Hence, the predicted label is two distance away from the actual label. But $r_i$ = 3 represents the second most similar query bucket. Thus, the impact of the performance of CORAL in this particular case is severe. On the other hand, the median value $\theta_{l_i}$ for $r_i$ = 3 bucket is 0.96. So if the predicted similarity score is far away from this value, it will be penalised by ATMSEL. So \name\ can clearly handle these cases better than CORAL.
 
\section{Conclusion and Future Work} In this paper, we develop a rank consistent method, ORDSIM for predicting query similarity. ORDSIM is architecture agnostic and can readily be implemented to extend multilayer perceptron. We experiment with an e-commerce dataset of 10 million queries and ORDSIM performs 42\% better than the best competing method. Currently, for representation learning of the e-commerce queries, we use BERT and spherical text embedding. Representing queries with spherical text embedding can capture the query similarity better than other embedding methods. Moreover, the hierarchical information of the categories for representing queries did not add necessary information so far. One reason can be -- categories are embedded in Poincaré space while queries are embedded in Euclidean space. In future, we want to use a kernel which transforms both the representations in a different space, and a carefully designed neural network can capture query similarity in that space.

\bibliographystyle{ACM-Reference-Format}
\bibliography{main}

\end{document}